# Theory-guided Auto-Encoder for Surrogate Construction and Inverse Modeling

Nanzhe Wang[a], Haibin Chang[a,*], and Dongxiao Zhang[b,*]

[a] BIC-ESAT, ERE, and SKLTCS, College of Engineering, Peking University, Beijing 100871, P. R. China
[b] School of Environmental Science and Engineering, Southern University of Science and Technology, Shenzhen 518055, P. R. China

[*] Corresponding author: E-mail address: changhaibin@pku.edu.cn (Haibin Chang); zhangdx@sustech.edu.cn (Dongxiao Zhang)

**Abstract**

A Theory-guided Auto-Encoder (TgAE) framework is proposed for surrogate construction and is further used for uncertainty quantification and inverse modeling tasks. The framework is built based on the Auto-Encoder (or Encoder-Decoder) architecture of convolutional neural network (CNN) via a theory-guided training process. In order to achieve the theory-guided training, the governing equations of the studied problems can be discretized and the finite difference scheme of the equations can be embedded into the training of CNN. The residual of the discretized governing equations as well as the data mismatch constitute the loss function of the TgAE. The trained TgAE can be used to construct a surrogate that approximates the relationship between the model parameters and responses with limited labeled data. In order to test the performance of the TgAE, several subsurface flow cases are introduced. The results show the satisfactory accuracy of the TgAE surrogate and efficiency of uncertainty quantification tasks can be improved with the TgAE surrogate. The TgAE also shows good extrapolation ability for cases with different correlation lengths and variances. Furthermore, the parameter inversion task has been implemented with the TgAE surrogate and satisfactory results can be obtained.

# 1 Introduction

Deep learning has been widely used for surrogate construction in various fields, such as hydrology (Mo et al., 2020; Mo et al., 2019), petroleum engineering (Jin et al., 2020; Tang et al., 2020), biomechanics (Liang et al., 2018), and so on. However, most of deep learning-based surrogates are purely data-driven, the domain knowledge such as scientific laws behind the studied problems is not considered during the surrogate construction process. Therefore, a large number of labeled training data are needed to guarantee the accuracy of the deep learning surrogates and some unreasonable predictions may be obtained from the data-driven model.

Recently, the many researchers have tried to combine the data-driven models and physics-based models by incorporating the physical constraints into the deep learning models. The most noticed work is the Physics-Informed Neural Network (PINN) proposed by Raissi et al. (2019). The residuals of the governing equations are added into the loss function of the neural network in the PINN framework. By minimizing the data mismatch as well as the residuals of the partial differential equations (PDEs), the network can honor not only the available data but also adhere to the governing equations of the physical problems. The PINN was used for solving the PDEs and identifying the parameters of the PDEs in their work. More recently, Wang et al. (2020) put forward a more general framework, Theory-guided Neural Network (TgNN), which can incorporate not only physical principles, but also some practical engineering theories. Furthermore, the TgNN has also been developed as a surrogate by Wang et al. (2021), which can be used for efficient uncertainty quantification of subsurface flow problems. Sun et al. (2020) applied the physical-constrained neural network to construct surrogate for fluid flows governed by Navier–Stokes equations, which was used for uncertainty propagation of fluid properties and domain geometry.

The works mentioned above are mainly based on fully-connected neural network architecture, and the physics-constrained strategy has also been extended to convolutional neural network architecture by some researchers. Zhu et al. (2019) proposed a physics-constrained deep learning surrogate which employs on the convolutional encoder-decoder neural network. The Sobel filter (Gao et al., 2010) was used in their work to approximate the spatial gradients, which can then be used to enforce the physical constraints on the convolutional architecture. Zheng et al. (2020) proposed a physics-informed semantic inpainting framework for heterogeneous geological field estimation, which employs a Wasserstein Generative Adversarial Network architecture and incorporates the physical laws. In their work, the Sobel filter was also used to calculate the spatial gradients needed for implementing the physical constraints. However, the time gradients cannot be obtained with the Sobel filter, so it would be difficult to implement physical constraints for dynamic problems with this scheme. Thus, the cases studied in Zhu et al. (2019) and Zheng et al. (2020) are both steady state problems.

In this work, a Theory-guided Auto-Encoder (TgAE) framework is proposed, which employs the convolutional Auto-Encoder network architecture. Without relying on the Sobel filter, in the TgAE, the governing equations of the studied problems are discretized with the finite difference method, which can be operated on the images output from the Auto-Encoder. And the residuals of the discretized governing equations can be added into the loss function of network to constrain the outputs from the Auto-Encoder with the physical laws. Therefore, the finite difference scheme for solving the PDEs can be embedded into the training process of network, and the trained TgAE can learn the scheme and predict solutions for cases with new model parameters. Several subsurface flow cases are introduced to test the performance of the TgAE surrogate. The extrapolation ability of the TgAE surrogate for different correlation lengths and variances of stochastic fields is also tested. The TgAE surrogate shows satisfactory accuracy and extrapolation ability, so the uncertainty quantification tasks can be implemented efficiently with the TgAE surrogate. Moreover, by combining with Iterative Ensemble Smoother (IES) (Chang et al., 2017; Chen & Oliver, 2013), the TgAE surrogate can be used for efficient inverse modeling, and the results show the effectiveness of the TgAE surrogate based IES for parameter inversion.

The rest of this paper is organized as follows. In section 2, the subsurface flow physical problems are introduced and inverse modeling task is presented. In section 3, the convolutional neural network is first introduced, then the Auto-Encoder based surrogate and the Theory-guided Auto-Encoder framework is introduced. The IES algorithm is also illustrated in section 3. In section 4, several subsurface flow cases are considered to test the performance of proposed methods. Finally, the discussions and conclusions are given in section 5.

## 2 Problem Formulation

In this work, the dynamic subsurface flow problems are considered with the following governing equation:

$$S_s \frac{\partial h(\mathbf{x},t)}{\partial t} - \nabla \cdot (K(\mathbf{x})\nabla h(\mathbf{x},t)) = 0 \qquad (1)$$

subjected to the following initial and boundary conditions:

$$h(\mathbf{x},t) = h_D(\mathbf{x}), \ \mathbf{x} \in \Gamma_D \qquad (2)$$

$$K(\mathbf{x})\nabla h(\mathbf{x}) \cdot \mathbf{n}(\mathbf{x}) = g(\mathbf{x}), \ \mathbf{x} \in \Gamma_N \qquad (3)$$

$$h(\mathbf{x},0) = h_0(\mathbf{x}) \qquad (4)$$

where $S_s$ [$L^{-1}$] denotes the specific storage; $h(\mathbf{x},t)$ [L] denotes the hydraulic head; $K(\mathbf{x})$ [$LT^{-1}$] denotes the hydraulic conductivity; $h_D(\mathbf{x})$ [L] denotes the prescribed head on Dirichlet boundary segments $\Gamma_D$; $g(\mathbf{x})$ [$LT^{-1}$] denotes the prescribed flux across Neumann boundary segments $\Gamma_N$; $\mathbf{n}(\mathbf{x})$ denotes an outward unit vector normal to the boundary; and $h_0(\mathbf{x})$ [L] denotes the hydraulic head at $t=0$, i.e., the initial conditions. With the model parameters such as hydraulic conductivity $K(\mathbf{x})$ being available and the boundary and initial conditions being specified, the problems can be solved with numerical simulators, which can be termed as *forward problems* and formulated as (Oliver et al., 2008):

$$g(\mathbf{m}) = \mathbf{d} \qquad (5)$$

where $\mathbf{m}$ denotes the general model parameters, and refer in particular to hydraulic conductivity $K$ in the introduced subsurface flow problems; $g$ denotes the forward model such as simulators; and $\mathbf{d}$ denotes the model response with model parameters $\mathbf{m}$. However, in most cases, the model parameters are not known specifically for the subsurface flow problems, because the direct observing of underground properties is difficult. Therefore, the unknown model parameters usually need to be estimated with the indirect observations of the model responses, which can be termed as *inverse problems* and represented by (Oliver et al., 2008):

$$\mathbf{d}^{obs} = g(\mathbf{m}) + \varepsilon \tag{6}$$

where $\mathbf{d}^{obs}$ denotes the indirect observations of model responses and $\varepsilon$ denotes the noises during the observation process.

## 3 Methodology

### 3.1 Deep Convolutional Neural Network (CNN)

Convolutional Neural Network (CNN) is a typical structure in deep learning (Lecun et al., 1998), which has excellent performance for image processing and has been widely used in various fields. Many variants of CNN have also been proposed to improve the performance of CNN, such as AlexNet (Krizhevsky et al., 2012), VGG (Simonyan & Zisserman, 2014), GoogLeNet (Szegedy et al., 2015), ResNet (He et al., 2016), and so on. Although there are all kinds of new variants, the core parts of CNNs are still the convolutional layers, which consist of a series of convolutional kernels or filters with learnable weights and biases. The convolutional kernels will slide across the width and height of the inputted images and compute dot products, so the kernels can extract the spatially local correlation information and produce the feature maps of the inputs. Moreover, the parameter sharing scheme of convolutional kernels can dramatically reduce the number of CNN parameters. Now consider a two-dimensional input image $\mathbf{I}$, and the outputted feature maps can be produced with a series of kernels $\mathbf{k}^r$ ($r=1,2,...,N_{out}$), where $N_{out}$ denotes the number of output channels. Then the $r$th channel or feature map can be calculated with:

$$H_{i,j}^r(\mathbf{I}) = \sigma(\sum_{u=1}^{k_m'}\sum_{v=1}^{k_n'} k_{u,v}^r I_{i+u, j+v}), (r=1,2,...,N_{out}) \tag{7}$$

where $H_{i,j}^r$ denotes the $r$th feature map value at location $(i,j)$ with input image $\mathbf{I}$; $\sigma(.)$

denotes the nonlinear activation function, such as Tanh, ReLU (Goodfellow et al., 2016), and Swish (Ramachandran et al., 2018); $k_{u,v}^r$ and $I_{i+u,j+v}$ denote the element in $\mathbf{k}^r$ and $\mathbf{I}$, respectively; and $k_m' \times k_n'$ denotes the size of the kernels. Then the outputs $\mathbf{H}$ with $N_{out}$ feature maps $H^r$ ($r=1,2,...,N_{out}$) can be produced after the convolution process with the kernels.

## 3.2 Auto-Encoder for surrogate modeling

Auto-Encoder is an important architecture of deep neural network, which consists of two parts: an encoder and a decoder (Goodfellow et al., 2016), as shown in **Figure 1**. The inputted images will be transformed into low-dimensional 'codes' with a series of convolutional layers of encoder, and the decoder can reconstruct the desired images by decoding the codes. Therefore, the convolutional Auto-Encoder can be trained to perform an image-to-image regression, which strategy has been used for surrogate modeling (Mo et al., 2019). The relationship between inputs and outputs to be approximated can be formulated as:

$$\Omega : \mathbf{X} \to \mathbf{Y} \tag{8}$$

where $\mathbf{X} \in \mathbf{R}^{D_x \times H \times W}$ and $\mathbf{Y} \in \mathbf{R}^{D_y \times H \times W}$ denote the input and output images, respectively; $\Omega$ denotes the mapping between inputs and outputs; and $H \times W$ denotes the image size $H \times W$, and $D_x$ ($D_y$) denotes the number of channels of inputs (outputs).

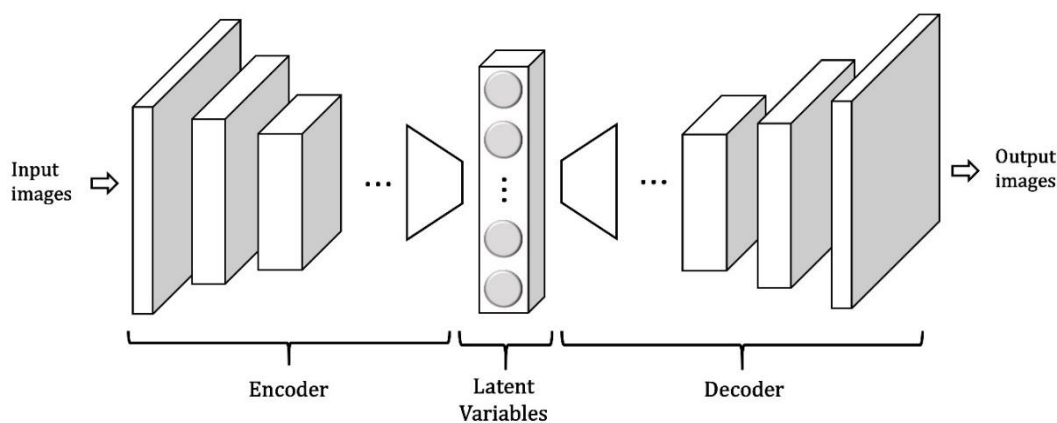

**Figure 1. The Architecture of Auto-Encoder.**

Then consider the dynamic subsurface flow problems introduced in section 2, which can be solved with finite difference methods. In order to improve the efficiency of the forward modeling, a convolutional Auto-Encoder based surrogate can be constructed for the subsurface problems. The approximated mapping between inputs and outputs by the surrogate can be expressed as:

$$f(\theta): \mathbf{X} \rightarrow \hat{\mathbf{Y}} \tag{9}$$

where $\theta$ denotes the parameters of the neural network and $\hat{\mathbf{Y}}$ denotes the estimated outputs via surrogate. Specifically, as shown in **Figure 2**, the inputs of the surrogate including the model parameters, hydraulic conductivity fields, which can be transformed into images, and the time matrixes, which are made up of the same elements. And the outputs of the surrogate are the distribution of hydraulic heads under specific parameter fields at a specific time. Therefore, the forward evaluation of the convolutional Auto-Encoder based surrogate can be expressed as:

$$f(K, T; \theta) = \hat{h}^{K,T} \tag{10}$$

where $K$ and $T$ denote the hydraulic conductivity fields and time matrixes, respectively; and $\hat{h}^{K,T}$ denotes the predicted hydraulic heads for conductivity field $K$ at time $T$.

A series of realizations can be solved with numerical simulators such as MODFLOW (Harbaugh, 2005), to constitute the training datasets for surrogate construction. Therefore, the surrogate modeling problem can be transformed into an image-to-image regression (Mo et al., 2019), and the Auto-Encoder based surrogate model can be trained with various optimization algorithms, such as stochastic gradient descent (SGD) (Bottou, 2010), and Adaptive Moment Estimation (Adam) (Kingma & Ba, 2015), to minimize the loss function:

$$L_{data}(\theta) = \frac{1}{N_r N_t} \sum_{i=1}^{N_r} \sum_{j=1}^{N_t} \left\| \hat{h}^{K_i, T_j} - h^{K_i, T_j} \right\|_2^2 = \frac{1}{N_r N_t} \sum_{i=1}^{N_r} \sum_{j=1}^{N_t} \left\| f(K_i, T_j; \theta) - h^{K_i, T_j} \right\|_2^2 \tag{11}$$

where $N_r$ and $N_t$ denote the total number of realizations solved to serve as training data and total number of time steps in each realization, respectively. After the training process, the trained Auto-Encoder can serve as a surrogate and make predictions of hydraulic heads for the cases with new hydraulic conductivity fields.

### 3.3 Theory-guided Auto-Encoder for surrogate modeling

To train a deep learning surrogate, a lot of data are usually needed to guarantee its accuracy. Moreover, the deep learning based surrogates are usually trained as black-boxes, and the physical knowledge is not involved during the surrogate construction process. In this work, a theory-guided Auto-Encoder is proposed for surrogate construction, which can be trained with limited training data and the scientific theories are considered. Consider the subsurface flow problem introduced in section 2, and the governing equation Eq.(1) can be discretized with the finite difference method:

$$S_s \frac{h_{i,j,n} - h_{i,j,n-1}}{\Delta t} - \frac{K_{i+1/2,j}(h_{i+1,j,n} - h_{i,j,n}) - K_{i-1/2,j}(h_{i,j,n} - h_{i-1,j,n})}{(\Delta x)^2}$$
$$- \frac{K_{i,j+1/2}(h_{i,j+1,n} - h_{i,j,n}) - K_{i,j-1/2}(h_{i,j,n} - h_{i,j-1,n})}{(\Delta y)^2} = 0 \quad (12)$$
$$(i = 1, 2, ..., H;\ j = 1, 2, ..., W;\ n = 1, 2, ..., N_t)$$

where $\Delta x, \Delta y, \Delta t$ denote the discretization interval in horizontal, vertical and time dimension, respectively; the subscript $i, j, n$ denote the indexes of discretized grids in horizontal, vertical and time dimension, respectively; and $K_{i+1/2,j}$ denotes the harmonic mean of hydraulic conductivities at grid $(i, j)$ and $(i+1, j)$, as shown following:

$$K_{i+1/2,j} = \frac{2 K_{i,j} \cdot K_{i+1,j}}{K_{i,j} + K_{i+1,j}} \quad (13)$$

In order to impose the physical constraints like governing equation into the training process, the residual of Eq.(12) calculated with the prediction can be minimized:

$$R(K,T;\theta) = S_s \frac{f(K,T;\theta)_{i,j} - f(K,T-\Delta t;\theta)_{i,j}}{\Delta t}$$
$$- \frac{K_{i+1/2,j}(f(K,T;\theta)_{i+1,j} - f(K,T;\theta)_{i,j}) - K_{i-1/2,j}(f(K,T;\theta)_{i,j} - f(K,T;\theta)_{i-1,j})}{(\Delta x)^2}$$
$$- \frac{K_{i,j+1/2}(f(K,T;\theta)_{i,j+1} - f(K,T;\theta)_{i,j}) - K_{i,j-1/2}(f(K,T;\theta)_{i,j} - f(K,T;\theta)_{i,j-1})}{(\Delta y)^2} \quad (14)$$
$$(i = 1, 2, ..., H;\ j = 1, 2, ..., W)$$

It can be seen that no labeled data of hydraulic heads are needed to calculate the residual $R(K,T;\theta)$ in Eq.(14), so a new concept can be proposed, i.e., *virtual realizations*, which are used to calculate the discretized PDE residuals and no numerical simulation runs are needed to solve those realizations. The concept of *virtual realizations* here is similar to the *collocation points* in the physical-informed neural network (Raissi et al., 2019), and no labeled values are needed at collocation points.

Therefore, the loss function to implement the physical constraints can be expressed as:

$$L_{PDE}(\theta) = \frac{1}{N_{vr}N_t} \sum_{i=1}^{N_{vr}} \sum_{j=1}^{N_t} \left\| R(K_i,T_j;\theta) \right\|_2^2 \tag{15}$$

where $N_{vr}$ denotes the total number of virtual realizations generated to impose the PDE constraints. Similarly, the Neumann boundary condition can also be constrained in the same way, and the discretized residual of Eq.(3) can be represented as:

$$R_{NB}(K,T;\theta) = \begin{bmatrix} \dfrac{K_{i-1/2,j}(f(K,T;\theta)_{i,j} - f(K,T;\theta)_{i-1,j})}{\Delta x} \\ \dfrac{K_{i,j-1/2}(f(K,T;\theta)_{i,j} - f(K,T;\theta)_{i,j-1})}{\Delta y} \end{bmatrix} \cdot \mathbf{n}(\mathbf{x}_{i,j}) - g_{i,j},$$

$$\mathbf{x}_{i,j} \in \Gamma_N \quad (i=1, 2, ..., H; j=1, 2, ..., W; t=1, 2, ..., N_t) \tag{16}$$

And the corresponding loss term can be expressed as:

$$L_{NB}(\theta) = \frac{1}{N_{vr}N_t} \sum_{i=1}^{N_{vr}} \sum_{j=1}^{N_t} \left\| R_{NB}(K_i,T_j;\theta) \right\|_2^2 \tag{17}$$

Moreover, the Dirichlet boundary conditions and the initial conditions can also be constrained by:

$$L_{DB}(\theta) = \frac{1}{N_{vr}N_t} \sum_{i=1}^{N_{vr}} \sum_{j=1}^{N_t} \left\| \hat{h}_{DB}^{K_i,T_j} - h_{DB}^{K_i,T_j} \right\|_2^2 = \frac{1}{N_{vr}N_t} \sum_{i=1}^{N_{vr}} \sum_{j=1}^{N_t} \left\| f_{DB}(K_i,T_j;\theta) - h_{DB}^{K_i,T_j} \right\|_2^2 \tag{18}$$

$$L_I(\theta) = \frac{1}{N_{vr}} \sum_{i=1}^{N_{vr}} \left\| \hat{h}^{K_i,T_0} - h^{K_i,T_0} \right\|_2^2 = \frac{1}{N_{vr}} \sum_{i=1}^{N_{vr}} \left\| f(K_i,T_0;\theta) - h^{K_i,T_0} \right\|_2^2 \tag{19}$$

where the subscript $DB$ of $\hat{h}_{DB}^{K_i,T_j}$ and $h_{DB}^{K_i,T_j}$ denotes the positions where they are set as Dirichlet boundary conditions on the images; and $T_0$ denotes the initial time matrix. Therefore, the final loss function can be expressed as:

$$L(\theta) = L_{data}(\theta) + L_{PDE}(\theta) + L_{NB}(\theta) + L_{DB}(\theta) + L_I(\theta) \tag{20}$$

Then the theory-guided Auto-Encoder (TgAE) surrogate $f(K,T;\theta)$ can be constructed by minimizing the loss function Eq.(20), which can not only honor the available labeled data, but can also adhere to the physical laws of the problems. While only the hydraulic heads at a single time step under a specific conductivity field can be obtained via the TgAE surrogate $f(K,T;\theta)$, and a circulation for all time steps is needed to finish a complete forward evaluation:

$$\hat{h}^K = g^{surr}(K) = \begin{bmatrix} f(K,T_1;\theta) \\ f(K,T_2;\theta) \\ \vdots \\ f(K,T_{N_t};\theta) \end{bmatrix} \tag{21}$$

where $g^{surr}$ denotes the constructed surrogate for the relationship Eq.(5) (between hydraulic conductivity and hydraulic heads in the subsurface flow cases); $\hat{h}^K$ denotes the estimated distribution of hydraulic heads under conductivity field $K$. The structure of the TgAE is presented in **Figure 2**.

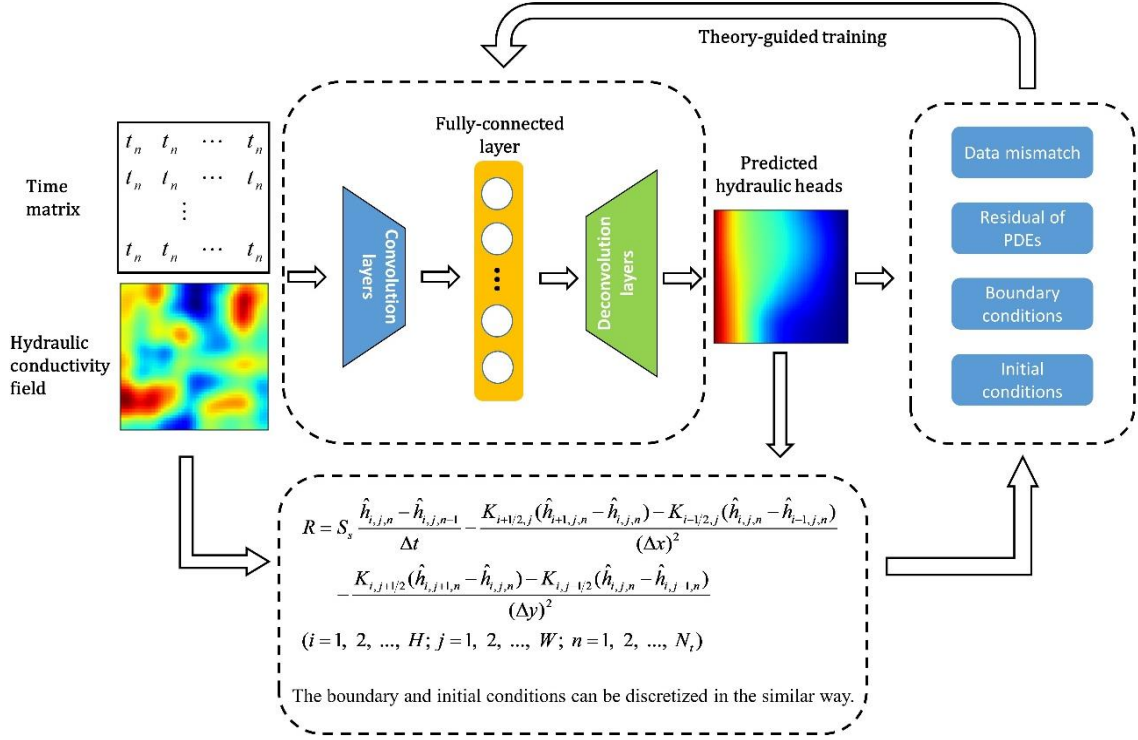

**Figure 2. The structure of TgAE.**

## 3.4 TgAE surrogate based iterative Ensemble smoother (IES) for inverse modeling

The iterative Ensemble smoother (IES) can be combined with the TgAE surrogate to improve the efficiency of inverse modeling. Under some reasonable assumptions, the inverse problem can be transformed into an optimization problem (Oliver et al., 2008):

$$O(\mathbf{m}) = \frac{1}{2}\left(\mathbf{d}^{obs} - g(\mathbf{m})\right)^T C_D^{-1}\left(\mathbf{d}^{obs} - g(\mathbf{m})\right) \\ + \frac{1}{2}\left(\mathbf{m} - \mathbf{m}^{pr}\right)^T C_M^{-1}\left(\mathbf{m} - \mathbf{m}^{pr}\right) \tag{22}$$

where $C_D$ and $C_M$ denote the covariance matrix of the observation error and model parameters, respectively; and $\mathbf{m}^{pr}$ denotes the prior estimation of model parameters. In order to minimize the objective function in Eq.(22), the Levenberg–Marquardt algorithm can be adopted to update the model parameters iteratively with the following scheme (Chang et al., 2017; Chen & Oliver, 2013):

$$\begin{aligned} \mathbf{m}_{l+1} &= \mathbf{m}_l - \left[(1+\lambda_l)C_M^{-1} + G_l^T C_D^{-1} G_l\right]^{-1} \left[C_M^{-1}(\mathbf{m}_l - \mathbf{m}^{pr}) + G_l^T C_D^{-1}(g(\mathbf{m}_l) - \mathbf{d}^{obs})\right] \\ &= \mathbf{m}_l - \frac{1}{1+\lambda_l}\left[C_M - C_M G_l^T \left((1+\lambda_l)C_D + G_l C_M G_l^T\right)^{-1} G_l C_M\right] C_M^{-1}(\mathbf{m}_l - \mathbf{m}^{pr}) \quad (23) \\ &\quad - C_M G_l^T \left((1+\lambda_l)C_D + G_l C_M G_l^T\right)^{-1} \left(g(\mathbf{m}_l) - \mathbf{d}^{obs}\right) \end{aligned}$$

where $l$ denotes the iteration number; $G_l$ denotes the sensitivity matrix of model response towards the model parameters at iteration step $l$; and $\lambda_l$ denotes the multiplier to the $C_M^{-1}$ to modify the Hessian matrix and reduce the impact of large data mismatch at earlier iteration stage (Chang et al., 2017; Chen & Oliver, 2013; Li et al., 2003). However, the calculation of sensitivity matrix $G_l$ is relatively difficult and tedious, and directly updating it with Eq.(23) is impractical. Therefore, the ensemble based methods provide an effective way to deal with those problems. In the ensemble based methods, a group of realizations would be updated in each iteration with Eq.(23), which can be represented by:

$$\begin{aligned} \mathbf{m}_{l+1,j} &= \mathbf{m}_{l,j} - \frac{1}{1+\lambda_l}\left[C_M - C_M G_{l,j}^T \left((1+\lambda_l)C_D + G_{l,j} C_M G_{l,j}^T\right)^{-1} G_{l,j} C_M\right] C_M^{-1}(\mathbf{m}_{l,j} - \mathbf{m}_j^{pr}) \\ &\quad - C_M G_{l,j}^T \left((1+\lambda_l)C_D + G_{l,j} C_M G_{l,j}^T\right)^{-1} \left(g(\mathbf{m}_{l,j}) - \mathbf{d}_j^{obs}\right), \quad j=1,\ldots,N_e \end{aligned} \quad (24)$$

where $N_e$ denotes the number of realizations in the ensemble; and subscript $j$ denotes the index of realizations. Then the iteration scheme can be modified by replacing $C_M$ with $C_{M_l}$, i.e., the covariance matrix at iteration step $l$, and replacing $G_{l,j}$ with $\bar{G}_l$, i.e., the average sensitivity matrix of all the realizations:

$$\begin{aligned} \mathbf{m}_{l+1,j} &= \mathbf{m}_{l,j} - \frac{1}{1+\lambda_l}\left[C_{M_l} - C_{M_l} \bar{G}_l^T \left((1+\lambda_l)C_D + \bar{G}_l C_{M_l} \bar{G}_l^T\right)^{-1} \bar{G}_l C_{M_l}\right] C_M^{-1}(\mathbf{m}_{l,j} - \mathbf{m}_j^{pr}) \\ &\quad - C_{M_l} \bar{G}_l^T \left((1+\lambda_l)C_D + \bar{G}_l C_{M_l} \bar{G}_l^T\right)^{-1} \left(g(\mathbf{m}_{l,j}) - \mathbf{d}_j^{obs}\right), \quad j=1,\ldots,N_e \end{aligned} \quad (25)$$

In order to avoid calculating the sensitivity matrixes directly, the following approximations are exploited (Chang et al., 2017; Le et al., 2016; Zhang, 2001):

$$C_{M_l} \bar{G}_l^T \approx C_{M_l D_l} \quad (26)$$

$$\bar{G}_l C_{M_l} \bar{G}_l^T \approx C_{D_l D_l} \quad (27)$$

where $C_{M_l D_l}$ denotes the cross covariance between the updated model parameters and the model response at iteration step $l$; and $C_{D_l D_l}$ denotes the covariance of the model response at iteration step $l$. Then the Eq.(25) can be rewritten as:

$$\begin{aligned} \mathbf{m}_{l+1,j} = \mathbf{m}_{l,j} &- \frac{1}{1+\lambda_l}\left[C_{M_l} - C_{M_l D_l}\left((1+\lambda_l)C_D + C_{D_l D_l}\right)^{-1}C_{D_l M_l}\right]C_M^{-1}\left(\mathbf{m}_{l,j} - \mathbf{m}_j^{pr}\right) \\ &- C_{M_l D_l}\left((1+\lambda_l)C_D + C_{D_l D_l}\right)^{-1}\left(g(\mathbf{m}_{l,j}) - \mathbf{d}_j^{obs}\right), \quad j=1,\ldots,N_e \end{aligned} \quad (28)$$

Since the forward modeling with simulation $g(\cdot)$ can be surrogated by the TgAE $g^{surr}(\cdot)$ to improve the efficiency of inversion, the Eq.(28) can be further rewritten as:

$$\begin{aligned} \mathbf{m}_{l+1,j} = \mathbf{m}_{l,j} &- \frac{1}{1+\lambda_l}\left[C_{M_l} - C_{M_l D_l}\left((1+\lambda_l)C_D + C_{D_l D_l}\right)^{-1}C_{D_l M_l}\right]C_M^{-1}\left(\mathbf{m}_{l,j} - \mathbf{m}_j^{pr}\right) \\ &- C_{M_l D_l}\left((1+\lambda_l)C_D + C_{D_l D_l}\right)^{-1}\left(g^{surr}(\mathbf{m}_{l,j}) - \mathbf{d}_j^{obs}\right), \quad j=1,\ldots,N_e \end{aligned} \quad (29)$$

where $\mathbf{m}$ denotes the general form of model parameters, and can be the formulated column vector with the hydraulic conductivity fields in the subsurface flow cases. In the parameters update process, the data mismatch between the measurements and the predictions based on estimated parameters is monitored and used to assess the accuracy of the parameters, which can be calculated with:

$$MIS(\mathbf{m}) = \frac{1}{N_e}\frac{1}{N_d}\sum_{j=1}^{N_e}\sum_{i=1}^{N_d}\left|\left(g^{surr}(\mathbf{m}_j)\right)_i - \mathbf{d}_{j,i}^{obs}\right|^2 \quad (30)$$

where $N_d$ denotes the total number of measurement data. The terminating criteria for iteration can be defined as:

(1) $MAX_{1\leq j\leq Ne; 1\leq i\leq n}\left|\mathbf{m}_{l+1,j}^i - \mathbf{m}_{l,j}^i\right| < \varepsilon_1$;

(2) $\dfrac{|MIS(\mathbf{m}_{l+1}) - MIS(\mathbf{m}_l)|}{\max(1,|MIS(\mathbf{m}_l)|)} < \varepsilon_2$;

(3) Iteration reaches the pre-given maximum iteration number $I_{max}$.

where $\varepsilon_1$ and $\varepsilon_2$ are the predefined limits of error; and $n$ denotes the dimension of the model parameter.

# 4 Case studies

Several subsurface flow cases are introduced to test the performance of the proposed TgAE surrogate in this section, which satisfy the governing equation Eq.(1) introduced in section 2. Specifically, the physical domain of the cases is a square with length of sides being $1020[L]$, where $[L]$ denotes the consistent length unit. The left and right sides of the domain are subjected to Dirichlet boundary conditions, with constant hydraulic heads $202[L]$ and $200$ $[L]$, respectively. The upper and lower boundaries are subjected to Neumann boundary conditions, being no flow boundaries in this work. And the domain, except for the left side, keeps the hydraulic heads being $200[L]$ at initial time step. The specific storage takes a constant value, $S_s = 0.0001\ [L^{-1}]$. The hydraulic conductivity fields can be generated with Karhunen–Loeve expansion (KLE) (Zhang & Lu, 2004). Moreover, the cases can be solved with MODFLOW to provide training labeled data (Harbaugh, 2005), and the domain is discretized into $51 \times 51$ grid blocks with $\Delta x = \Delta y = 20[L]$. The total simulation time span is $10[T]$, which is divided into 50 time steps with time interval $\Delta t = 0.2[T]$.

## 4.1 Surrogate modeling and uncertainty quantification

In this subsection, the TgAE surrogate is constructed for the subsurface flow case and then used for uncertainty quantification of model responses towards stochastic hydraulic conductivity fields. In this case, the stochastic hydraulic conductivity field $K(\mathbf{x})$ is assumed to be lognormal, and the mean and variance of $\ln K(\mathbf{x})$ are set to be 0 and 1.0, respectively, i.e., $\langle \ln K \rangle = 0$ and $\sigma_{\ln K}^2 = 1.0$. The covariance function of $\ln K(\mathbf{x})$ is assumed to be exponential, as shown below:

$$C_{\ln K}(\mathbf{x}_1, \mathbf{x}_2) = \sigma_{\ln K}^2 \exp\left\{-\left[\left(\frac{|x_1 - x_2|}{\eta_x}\right) + \left(\frac{|y_1 - y_2|}{\eta_y}\right)\right]\right\} \tag{31}$$

where $\eta_x$ and $\eta_y$ denote the correlation length of the random field, which take the value of 408 [L] in this case; and $\mathbf{x}_i = (x_i, y_i)$ denotes the coordinate of a grid block. Then the realizations can be generated with KLE. More details of KLE can be found in Zhang and Lu (2004). In this case, 80% energy of the stochastic fields is preserved, which leads to 20 truncated terms in KLE. 30 realizations are generated with KLE and solved with the MODFLOW simulator to provide labeled training dataset. Moreover, 150 virtual realizations are also generated with KLE and used to impose the physical constraints. It is worth noting that the virtual realizations do not need to be solved with the simulator. 200 new test realizations are generated with KLE and solved with MODFLOW to constitute the testing dataset and to verify the accuracy of the trained surrogate.

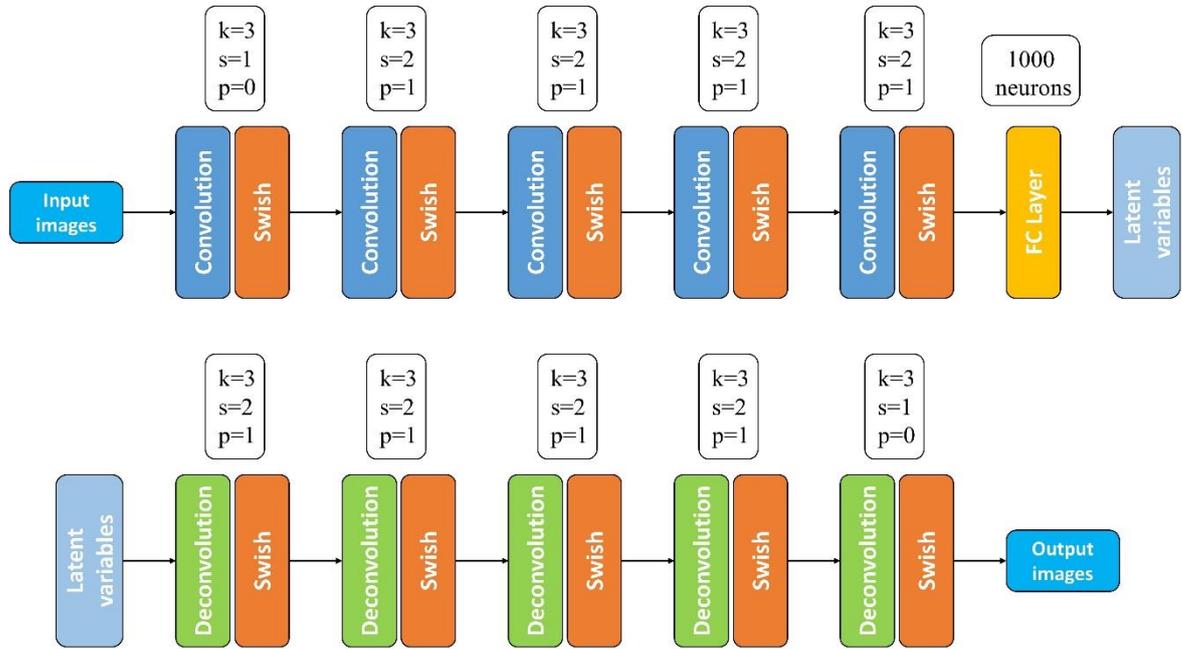

**Figure 3. The Auto-Encoder architecture used in this work.**

Then the TgAE can be constructed for surrogate modeling. Because the domain is discretized into $51\times 51$ grid blocks, the resolution of input and output images is $51\times 51$. The inputs consist of 2 channels, i.e., hydraulic conductivity field image and time matrix, and the outputs have 1 channel, i.e., hydraulic heads image. And the detailed network structure of TgAE is shown in **Figure 3**. The activation function, Swish, is used after each convolution and deconvolution layer. The Adam algorithm is then used to train the network with learning rate of 0.001 for 1000 epochs. And it takes approximately 0.824 h (2967.623 s) to train the TgAE surrogate on an NVIDIA TITAN RTX GPU card.

### *4.1.1 TgAE surrogate evaluation*

The constructed surrogate is evaluated in this subsection. The solutions of the 200 test realizations are predicted with the trained TgAE surrogate, and three realizations are randomly sampled as shown in **Figure 4**. The hydraulic heads of the three realizations obtained from MODFLOW and the TgAE surrogate are shown in **Figure 5**, **Figure 6**, and **Figure 7**. It can be seen that the predictions from the TgAE surrogate are basically consistent with the references. And the hydraulic heads of three points in the studied domain for 200 test realizations solved with MODFLOW and predicted with the TgAE surrogate are shown in **Figure 8**, which shows the accuracy of the surrogate.

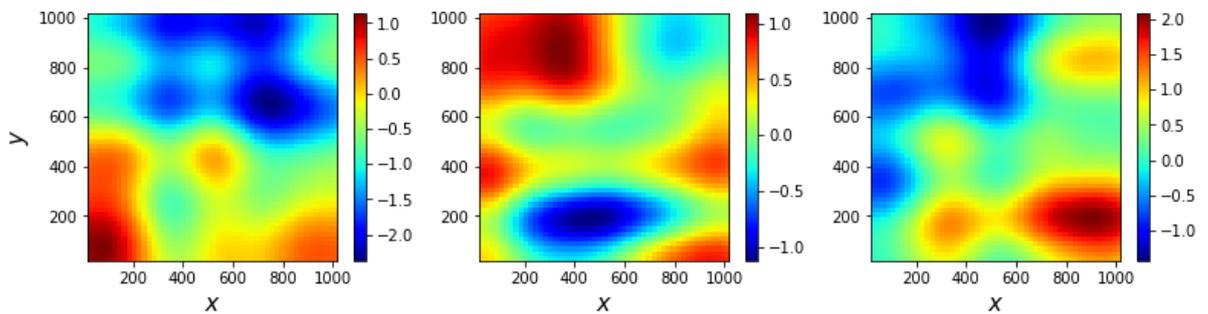

**Figure 4. Three randomly sampled realizations for testing the TgAE surrogate.**

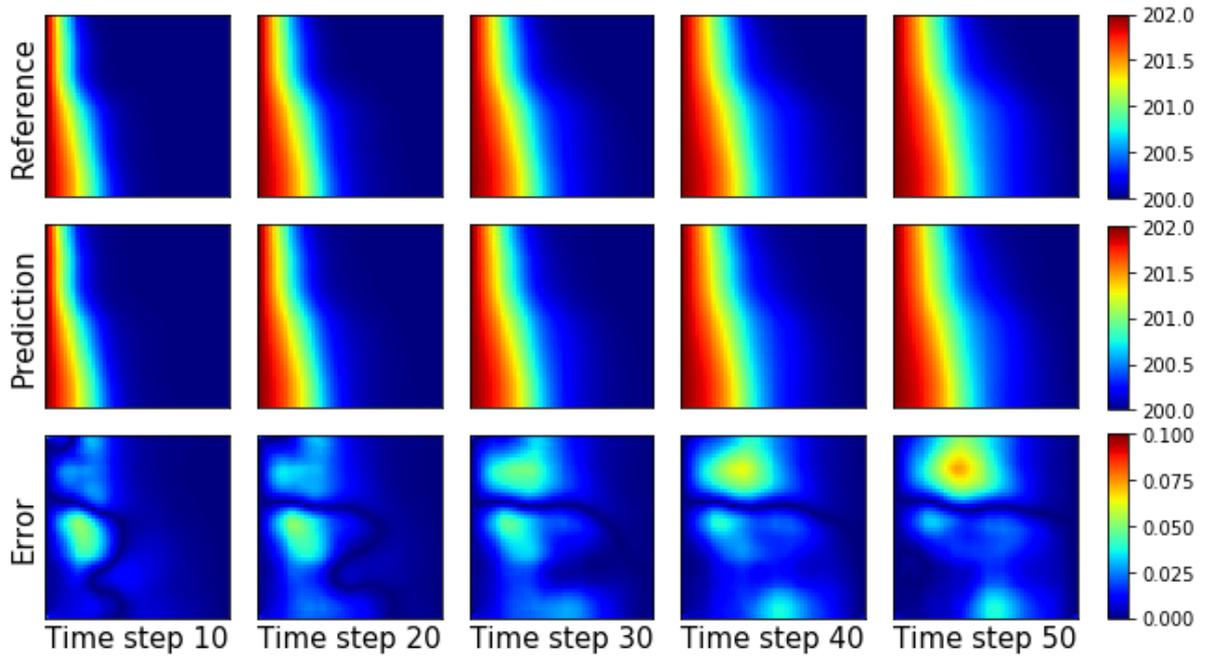

**Figure 5. The hydraulic heads obtained from MODFLOW and the TgAE surrogate for realization 1.**

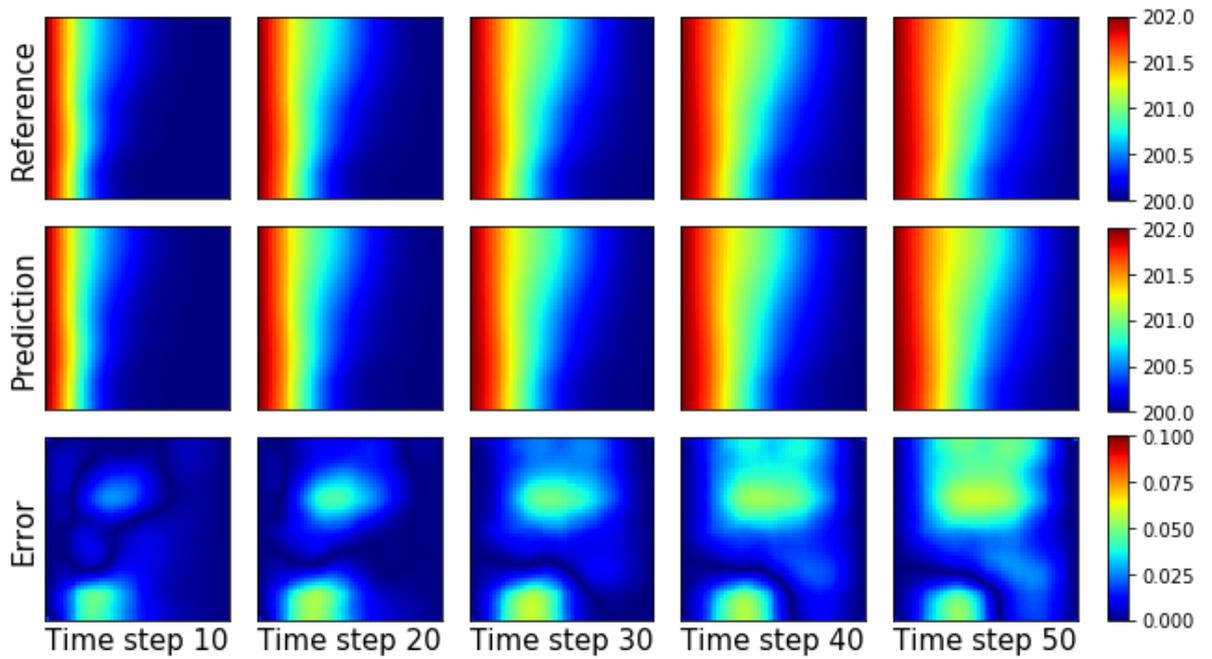

**Figure 6. The hydraulic heads obtained from MODFLOW and the TgAE surrogate for realization 2.**

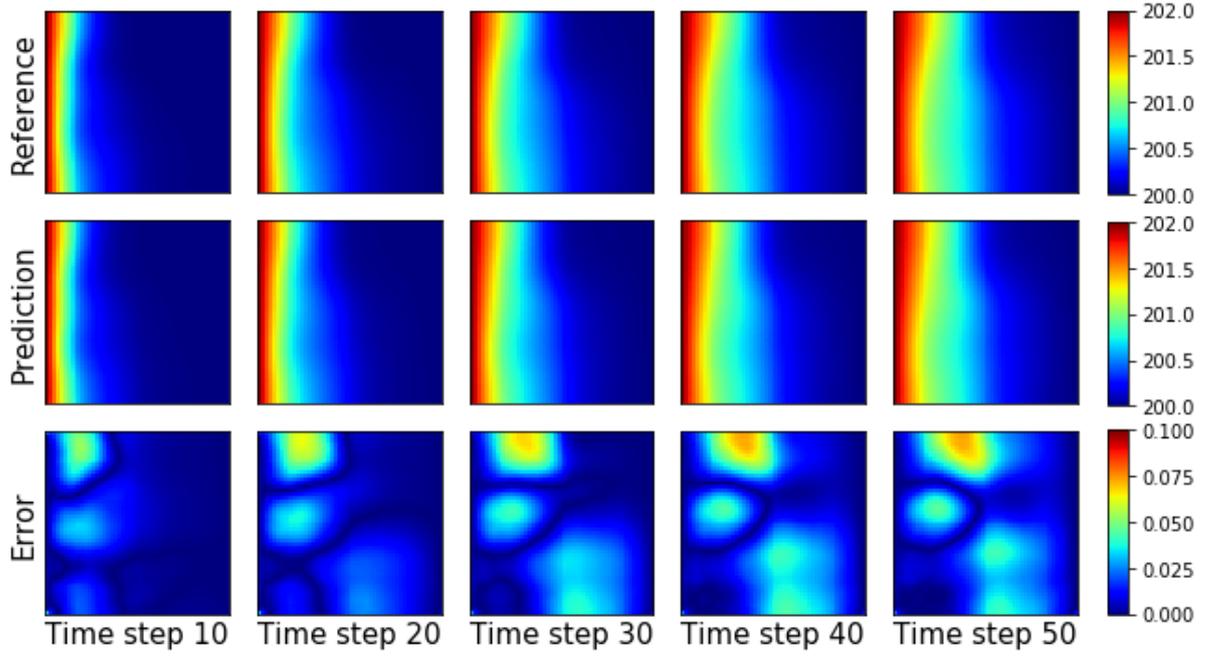

**Figure 7. The hydraulic heads obtained from MODFLOW and the TgAE surrogate for realization 3.**

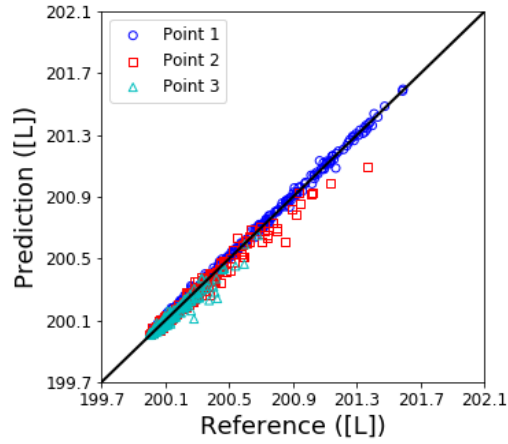

**Figure 8. Correlation between predictions from the TgAE surrogate and the reference from MODFLOW for three points of 200 testing realizations (point 1: x=200[L], y=200[L], t=2[T]; point 2: x=520[L], y=520[L], t=5[T]; point 3: x=200[L], y=800[L], t=8[T]).**

Moreover, in order to test the performance of the TgAE surrogate quantitatively, two evaluation metrics are introduced, i.e., the relative $L_2$ error and $R^2$ score, as shown in Eq.(32) and Eq.(33), respectively:

$$L_2 = \frac{\|h_{pred} - h_{ref}\|_2}{\|h_{ref}\|_2} \tag{32}$$

$$R^2 = 1 - \frac{\sum_{n=1}^{N_{cell}}(h_{pred,n} - h_{ref,n})^2}{\sum_{n=1}^{N_{cell}}(h_{ref,n} - \bar{h}_{ref})^2} \tag{33}$$

where $h_{ref}$ and $h_{pred}$ denote the reference value solved by MODFLOW and the prediction from the surrogate, respectively; $\|\cdot\|_2$ denotes the standard Euclidean norm; $N_{cell}$ denotes the number of blocks needed to be predicted; $h_{ref,n}$ and $h_{pred,n}$ denote the reference value solved by MODFLOW and the prediction from the surrogate at the $n^{th}$ block, respectively; and $\bar{h}_{ref}$ denotes the mean of $h_{ref,n}$. The histogram of the relative $L_2$ error and $R^2$ score for the 200 test realizations are shown in **Figure 9**. The results show the satisfactory accuracy of the trained TgAE surrogate.

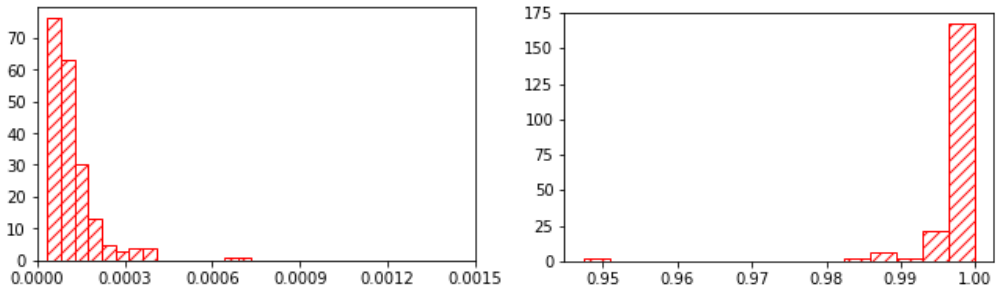

Figure 9. Histogram of the relative $L_2$ error and $R^2$ score for the 200 test realizations.

*4.1.2 Uncertainty quantification based on TgAE surrogate*

The constructed TgAE surrogate is used for uncertainty quantification in this subsection. The Monte Carlo (MC) method is used as benchmark for comparison. In the MC method, 10,000 realizations are generated by adopting KLE with 96% energy preserved, which are then solved with MODFLOW and used to calculate the statistical moments of the model responses. The TgAE surrogate is then used to predict the model responses for 10,000 realizations, which are truncated with 80% energy retained in KLE. The mean, variance, and Probability Density

Function (PDF) of the model responses can be estimated with the predictions from the surrogate. The uncertainty quantification results of this case is shown in **Figure 10**, **Figure 11**, and **Figure 12**, and the relative $L_2$ error and $R^2$ score of estimated mean and variance are shown in **Table 1**. It can be seen that the TgAE surrogate can be used to estimate the statistical moments of model responses towards stochastic inputs accurately. And it is worth noting that the prediction for 10,000 different realizations only takes 80.70s with the TgAE surrogate. However, approximately 1.75h (6,304.30s) are needed to solve 10,000 realizations with MODFLOW simulator, which shows the improvement of efficiency for uncertainty quantification with the TgAE surrogate.

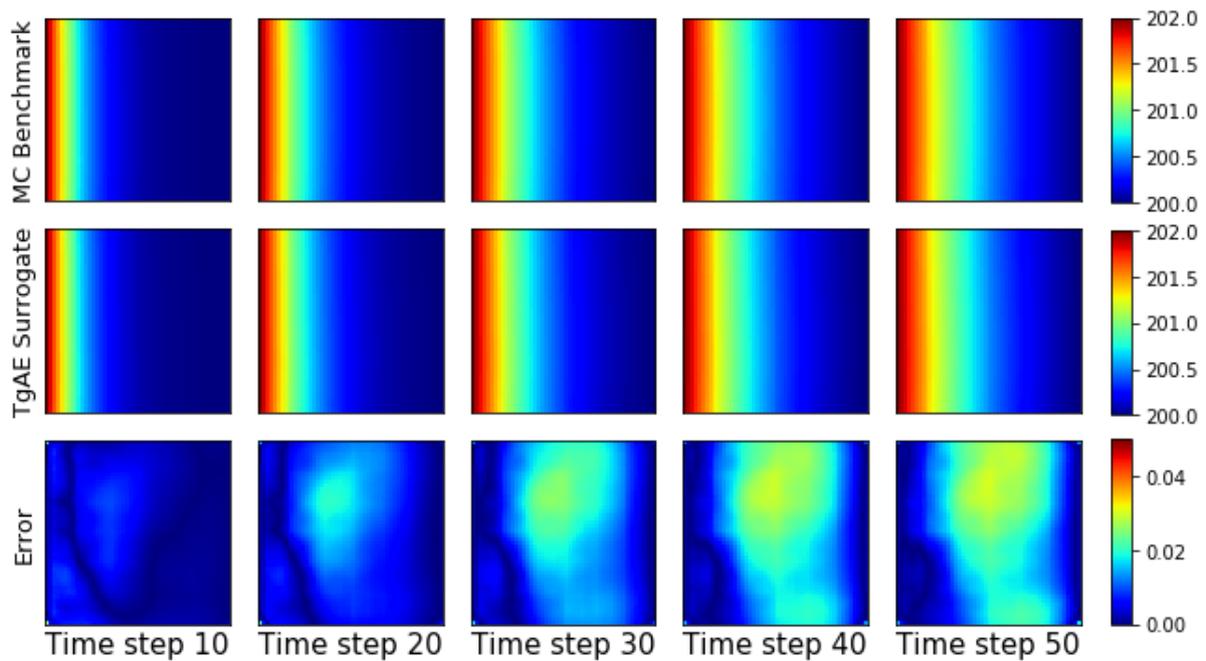

Figure 10. The mean of hydraulic head estimated with MC method and the TgAE surrogate.

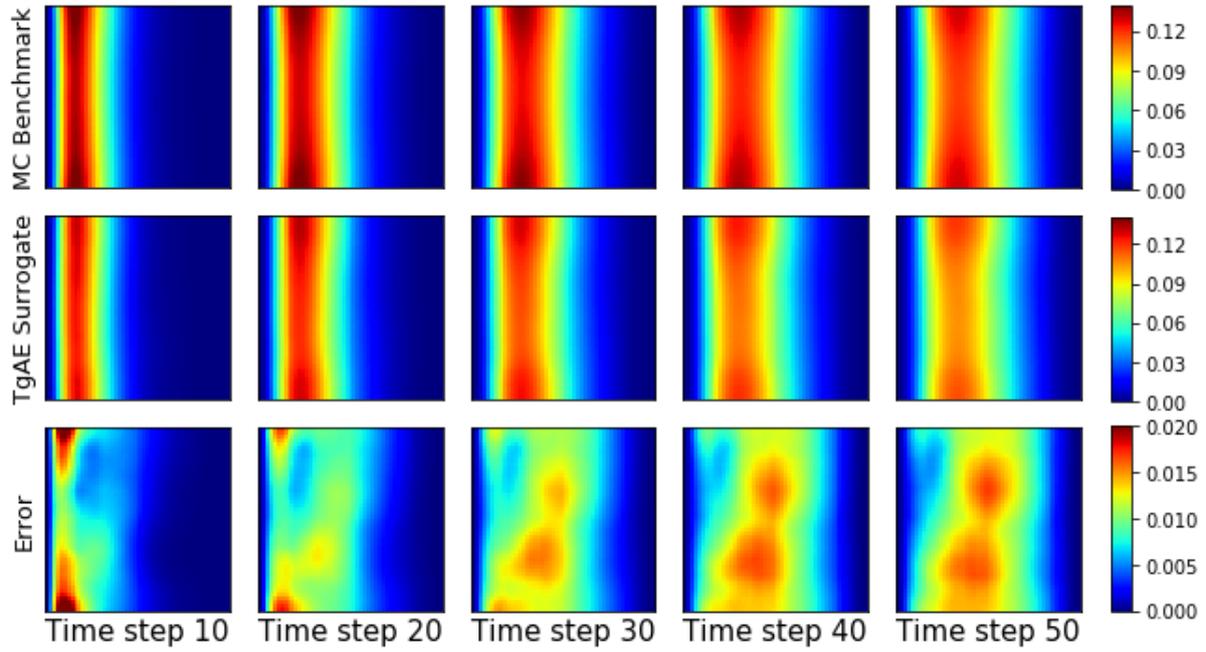

Figure 11. The variance of hydraulic head estimated with MC method and the TgAE surrogate.

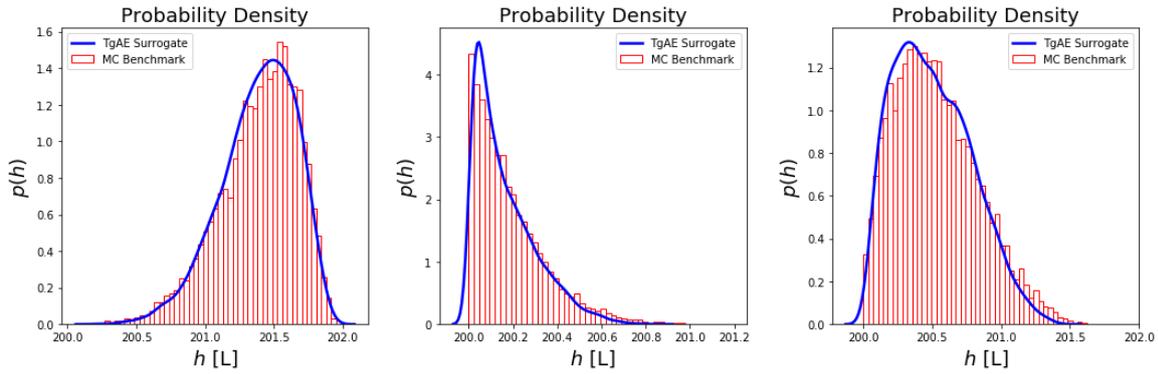

Figure 12. The PDF of hydraulic head at three points estimated with MC method and the TgAE surrogate (point 1: x=140[L], y=780[L], t=5.2[T]; point 2: x=780[L], y=780[L], t=9.2[T]; point 3: x=520[L], y=520[L], t=9.2[T]).

Table 1. The relative $L_2$ error and $R^2$ score of estimated mean and variance.

|  | relative $L_2$ error | $R^2$ score |
|---|---|---|
| **Mean** | 6.009e-05 | 9.996e-01 |
| **Variance** | 1.209e-01 | 9.696e-01 |

### 4.1.3 The effect of virtual realizations

In our previous work, the effect of labeled data and collocation points towards the accuracy of TgNN surrogate is studied (Wang et al., 2021). Similarly, the effect of virtual realizations number is tested in this subsection. The TgAE surrogates are trained with 30 solved realizations, which are used as labeled training data, and a different number of virtual realizations to impose the physical constraints. The correlation of reference values and predictions from different TgAE surrogates for 200 test realizations are shown in **Figure 13**. Moreover, the relative $L_2$ error and $R^2$ score of estimated statistical moments with different trained surrogates are shown in **Table 2**, and the corresponding training time for each surrogate is also listed in **Table 2**. It is obvious that the TgAE surrogate gets more accurate as the number of virtual realizations increases, and the corresponding uncertainty quantification results also get more accurate, which is intuitively correct. It also takes longer to train the TgAE surrogate as the number of virtual realizations increases.

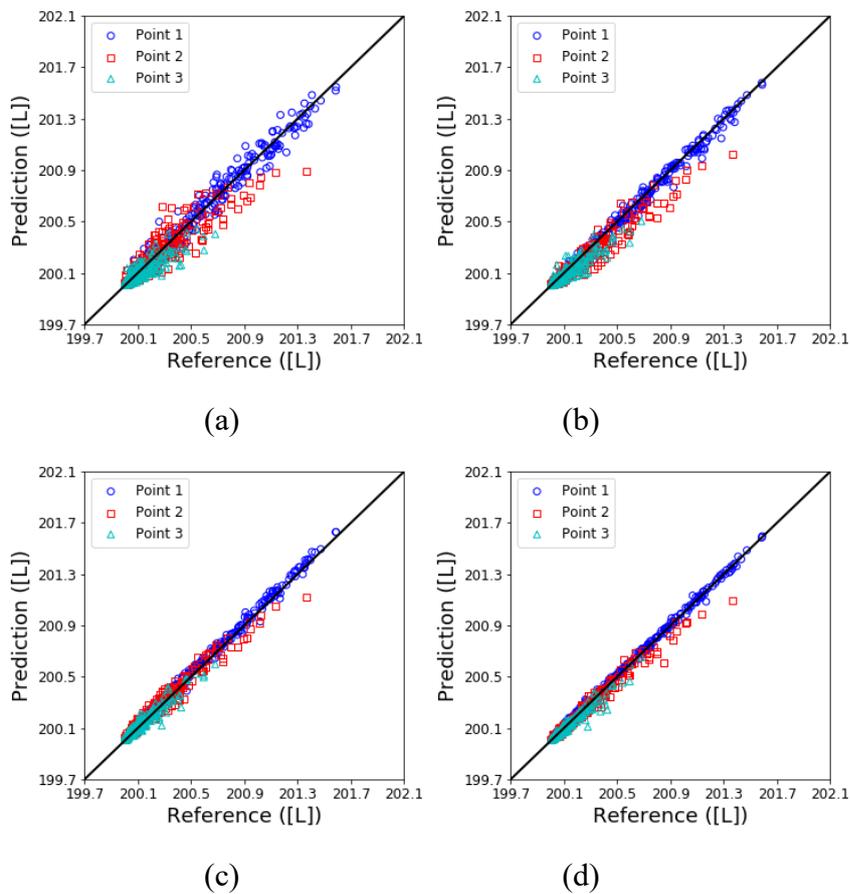

(a)  (b)

(c)  (d)

**Figure 13. Correlation of reference from MODFLOW and predictions from TgAE surrogates trained**

Table 2. The relative $L_2$ error and $R^2$ score of estimated statistical moments with different trained surrogates.

| Number of virtual realizations | Relative $L_2$ error | | $R^2$ score | | Training time(s) |
|---|---|---|---|---|---|
| | Mean | Variance | Mean | Variance | |
| Nv=0 | 8.900e-05 | 2.765e-01 | 9.990e-01 | 8.413e-01 | 1695.970 |
| Nv=50 | 1.541e-04 | 1.518e-01 | 9.971e-01 | 9.522e-01 | 1930.345 |
| Nv=100 | 6.043e-05 | 1.569e-01 | 9.996e-01 | 9.489e-01 | 2389.938 |
| Nv=150 | 6.009e-05 | 1.209e-01 | 9.996e-01 | 9.696e-01 | 2967.623 |

### 4.2 Extrapolation of TgAE surrogate for correlation length and variance

In this subsection, the extrapolating ability of the trained TgAE surrogate for different correlation lengths and variances of the stochastic field is tested. The TgAE surrogate is first trained with realizations of specific variance and correlation length, and then used for uncertainty quantification of scenarios with different variances and correlation lengths.

In this case, the training realizations are generated by using KLE with $\sigma^2_{\ln K}=1.0$ and $\eta_x=\eta_y=306[L]$. 90 realizations are solved with MODFLOW simulator serving as labeled dataset and 500 virtual realizations are generated to impose the physical constraints. It takes 0.97h (3506.220s) to train the surrogate on the NVIDIA TITAN RTX GPU card. The correlations of hydraulic heads at 3 different points from the simulator and TgAE surrogate for 200 test realizations are shown in **Figure 14**. Then the trained TgAE surrogate can be tested to predict the solutions for cases with different variances and correlation lengths. The testing results for different cases are shown in **Figure 15**. It can be seen that the trained TgAE surrogate can predict solutions for cases with different statistical information from the training datasets. Therefore, the trained TgAE surrogate can then be used for uncertainty quantification of cases with different correlation length and variance settings.

First, in order to test the extrapolation capacity for variance of the random field, the TgAE surrogate is used to quantify the model response uncertainty for cases with different variances and same correlation length ($\eta_x = \eta_y = 306[L]$). The benchmark for each case is also obtained from MC method with 10,000 realizations solved with MODFLOW, and the TgAE surrogate is then used to predict solutions for the 10,000 realizations of each case. The relative $L_2$ error and $R^2$ score of estimated means and variances for each case are listed in **Table 3**. It can be seen that the TgAE surrogate can perform well for cases with different variances. Second, the extrapolation capacity for correlation length of the TgAE surrogate is tested. The cases with different correlation lengths and same variance ($\sigma_{\ln K}^2 = 1.0$) are considered. The relative $L_2$ error and $R^2$ score of estimated statistical moments for each case are shown in **Table 4**, which shows the capability of the trained TgAE surrogate to extrapolate to different correlation lengths of stochastic fields. Therefore, there is no need to retrain the surrogate while studying cases with different statistical information, which helps to save a lot of computational cost. Moreover, it can be seen that only about 80s are needed to perform 10,000 forward evaluations for each case, and about 6000s are needed with MODFLOW simulator, which shows the efficiency of the TgAE surrogate based uncertainty quantification.

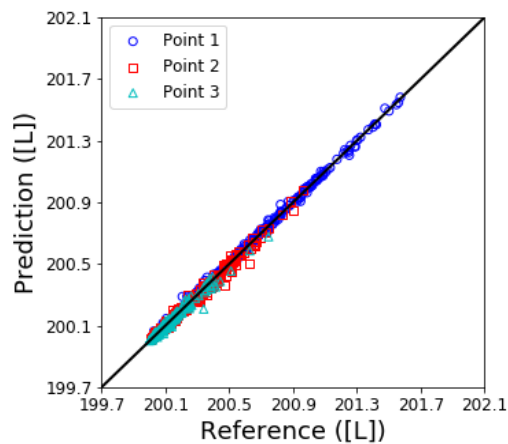

**Figure 14. Correlation between reference from MODFLOW and predictions from the TgAE surrogate for three points (point 1: x=200[L], y=200[L], t=2[T]; point 2: x=520[L], y=520[L], t=5[T]; point 3: x=200[L], y=800[L], t=8[T]) of the case with $\sigma_{\ln K}^2 = 1.0$ and $\eta_x = \eta_y = 306[L]$.**

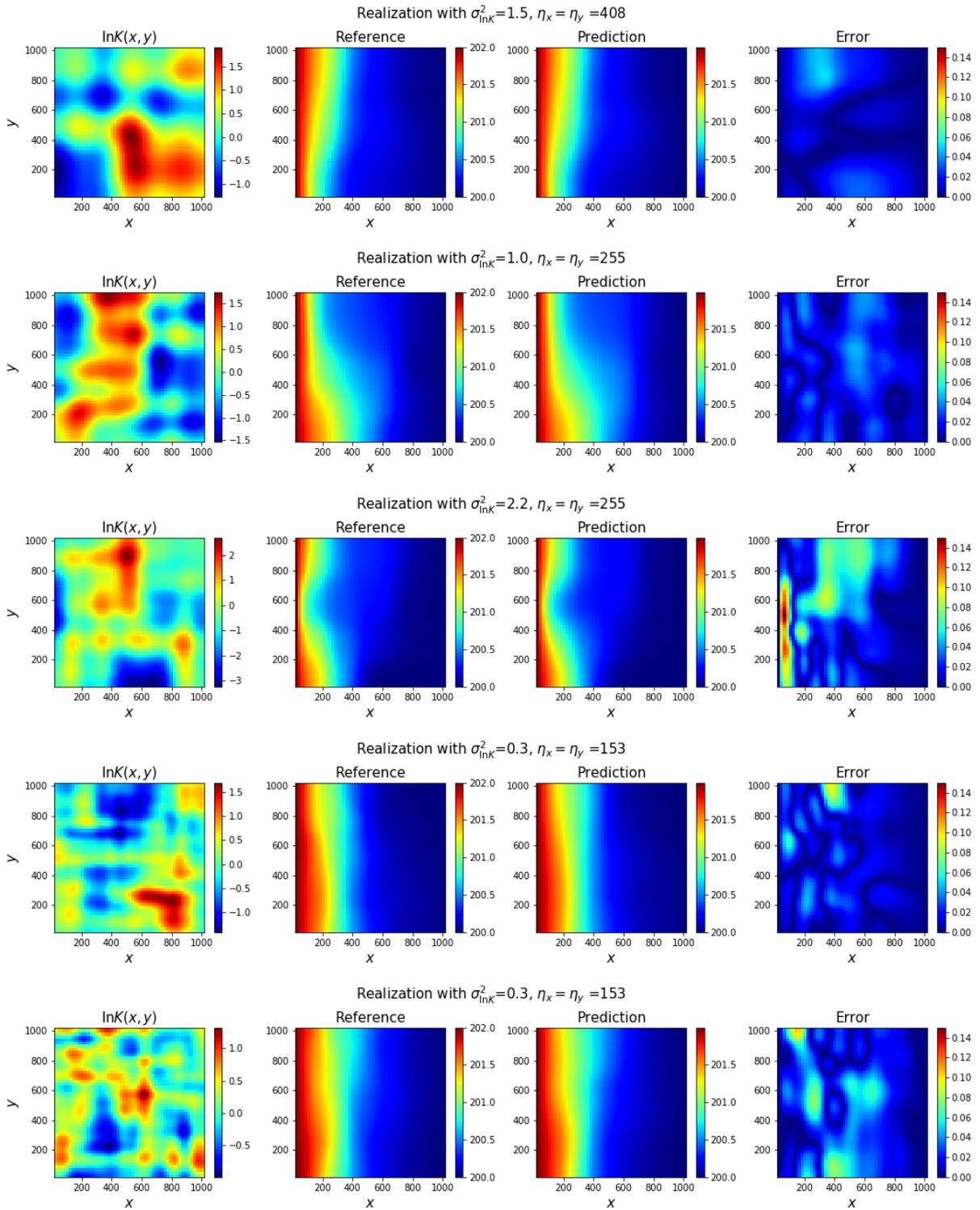

**Figure 15. The predictions of the TgAE surrogate for cases with different correlation lengths and variances.**

Table 3. The relative $L_2$ error and $R^2$ score of estimated means and variances for cases with different variances.

| Variance of random fields | Mean | | Variance | | Consumed Time with TgAE surrogate (s) | Consumed Time with MODFLOW (s) |
|---|---|---|---|---|---|---|
| | relative $L_2$ error | $R^2$ score | relative $L_2$ error | $R^2$ score | | |
| $\sigma^2_{lnK}$=0.3 | 1.5595e-05 | 9.9997e-01 | 5.2503e-02 | 9.9480e-01 | 82.5128 | 5922.391 |
| $\sigma^2_{lnK}$=1.0 | 2.4520e-05 | 9.9993e-01 | 6.5509e-02 | 9.9137e-01 | 84.0494 | 6038.978 |
| $\sigma^2_{lnK}$=1.5 | 4.7324e-05 | 9.9972e-01 | 7.3506e-02 | 9.8865e-01 | 86.4104 | 6064.349 |
| $\sigma^2_{lnK}$=2.0 | 6.8265e-05 | 9.9941e-01 | 8.0248e-02 | 9.8592e-01 | 85.3544 | 6028.007 |
| $\sigma^2_{lnK}$=2.2 | 7.6118e-05 | 9.9927e-01 | 8.2686e-02 | 9.8482e-01 | 82.6857 | 6134.306 |

Table 4. The relative $L_2$ error and $R^2$ score of estimated means and variances for cases with different correlation lengths.

| Correlation length of random fields | Mean | | Variance | | Consumed Time with TgAE surrogate (s) | Consumed Time with MODFLOW (s) |
|---|---|---|---|---|---|---|
| | relative $L_2$ error | $R^2$ score | relative $L_2$ error | $R^2$ score | | |
| $\eta_x=\eta_y$=510 | 2.1849e-05 | 9.9994e-01 | 7.7677e-02 | 9.8736e-01 | 84.1151 | 5923.579 |
| $\eta_x=\eta_y$=306 | 2.4520e-05 | 9.9993e-01 | 6.5509e-02 | 9.9137e-01 | 84.0494 | 6038.978 |
| $\eta_x=\eta_y$=255 | 2.1051e-05 | 9.9995e-01 | 4.8754e-02 | 9.9528e-01 | 81.8831 | 5869.759 |
| $\eta_x=\eta_y$=153 | 2.2564e-05 | 9.9994e-01 | 4.0797e-02 | 9.9693e-01 | 82.4004 | 5932.957 |

### 4.3 Surrogate modeling for non-Gaussian fields

In this subsection, the TgAE surrogate is tested with non-Gaussian fields. The stochastic fields generated with KLE are Gaussian, which can be transformed into non-Gaussian fields with a superposition of polynomial expansion (Zhang et al., 2011), i.e.,

$$\ln K(\mathbf{x}) = U_0(\mathbf{x}) + U_1(\mathbf{x})\gamma(\mathbf{x}) + U_2(\mathbf{x})\left(\gamma^2(\mathbf{x})-1\right) + U_3(\mathbf{x})\left(\gamma^3(\mathbf{x})-3\gamma(\mathbf{x})\right) + \cdots \qquad (34)$$

where $U_i(\mathbf{x})$ denotes the deterministic coefficients; $\gamma(\mathbf{x})$ denotes the Gaussian random fields, which can be generated with KLE, i.e.,

$$\gamma(\mathbf{x}) = \bar{\gamma}(\mathbf{x}) + \sum_{i=1}^{\infty} \sqrt{\lambda_i} f_i(\mathbf{x}) \xi_i \qquad (35)$$

where $\bar{\gamma}(\mathbf{x})$ denotes the mean of the random field; $\lambda_i$ and $f_i$ denote the eigenvalues and eigenfunctions of covariance function, respectively; and $\xi_i$ denotes the independent orthogonal Gaussian random variables with zero mean and unit variance. The Eq.(35) can be further truncated with finite terms according to the decay of eigenvalues $\lambda_i$, i.e.,

$$\gamma(\mathbf{x}) = \bar{\gamma}(\mathbf{x}) + \sum_{i=1}^{N} \sqrt{\lambda_i} f_i(\mathbf{x}) \xi_i \qquad (36)$$

where $N$ denotes the total retained terms in the expansion.

In this case, a 3$^{rd}$ order polynomial expansion is used to represent the non-Gaussian random fields, with $U_0=0$, $U_1=0.5$, and $U_2=0.3$. This transformation is illustrated with an example. Consider a Gaussian field generated with KLE shown in **Figure 16(a)**, and the histogram of the values at each grid block is shown in **Figure 16(b)**, which is a nearly Gaussian distribution. The transformed random field is shown in **Figure 17(a)**, and the distribution of the field values is left skewness as shown in **Figure 17(b)**. Therefore, the generated Gaussian realizations with KLE can be transformed into non-Gaussian fields with the preset polynomial expansion. The TgAE surrogate can then be constructed for the non-Gaussian realizations. Similarly, in this non-Gaussian case, 30 realizations are solved with the simulator and 250 virtual realizations are generated to impose the physical constraints. The surrogate is trained with Adam algorithm with a learning rate of 0.001, and it takes about 0.703h (2530.91s) for 300 epochs on the NVIDIA TITAN RTX GPU card. Three sampled realizations are shown in **Figure 18**, and the prediction results for the three realizations are presented in **Figure 19**, **Figure 20**, and **Figure 21**, which show the satisfactory accuracy of the constructed TgAE surrogate for the non-Gaussian fields.

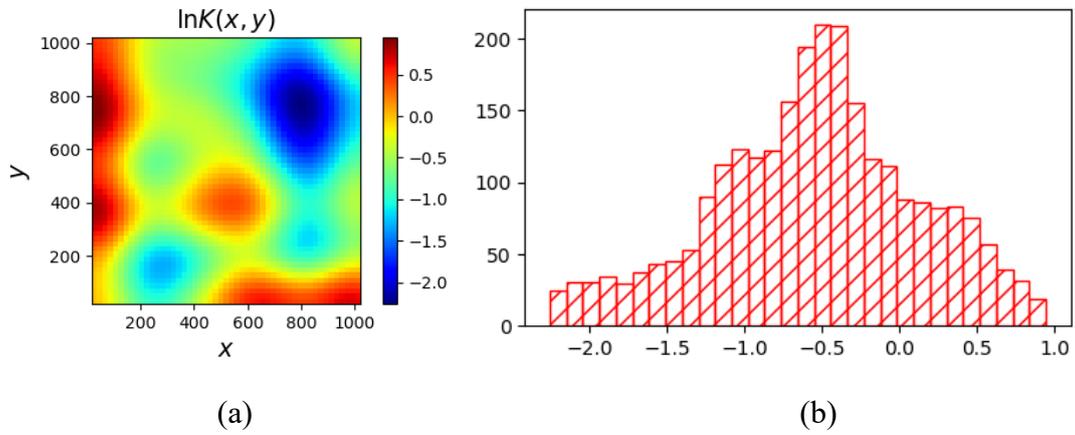

(a)          (b)

**Figure 16. Gaussian field generated with KLE (a) and**

**the histogram of the values at each grid block (b).**

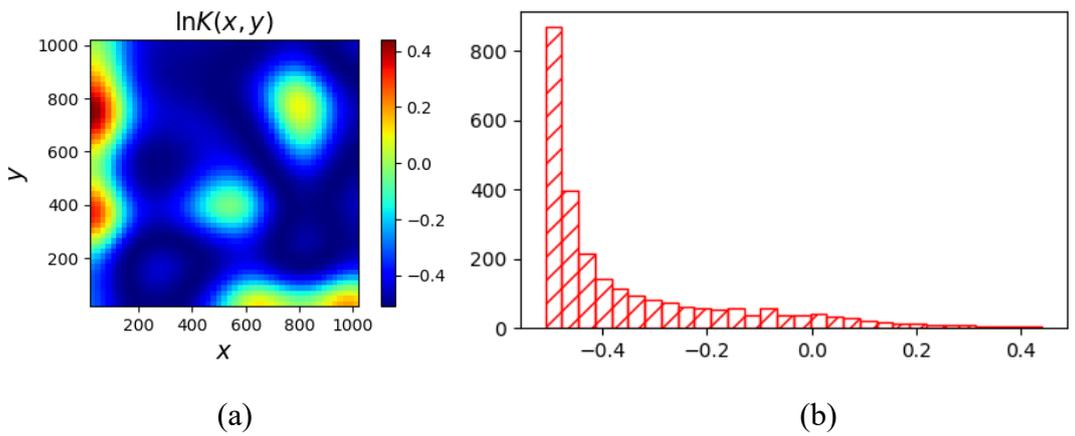

(a)          (b)

**Figure 17. Non-Gaussian field transformed with polynomial expansion (a) and the histogram of the**

**values at each grid block (b).**

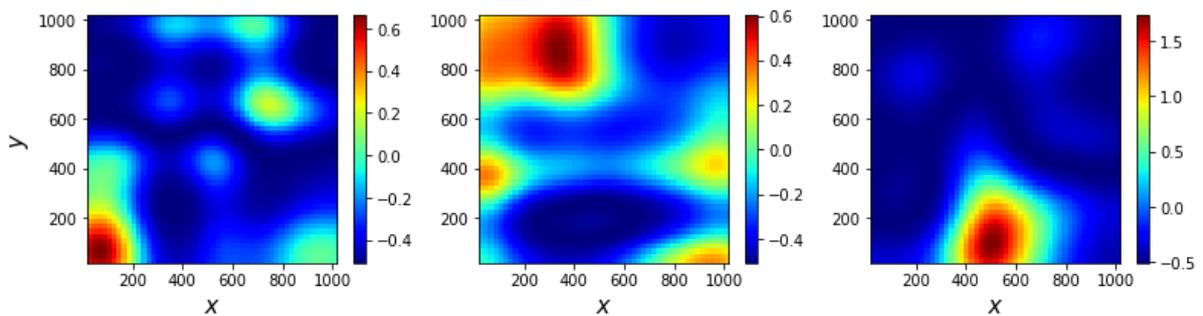

**Figure 18. Three randomly sampled non-Gaussian realizations for testing the TgAE surrogate.**

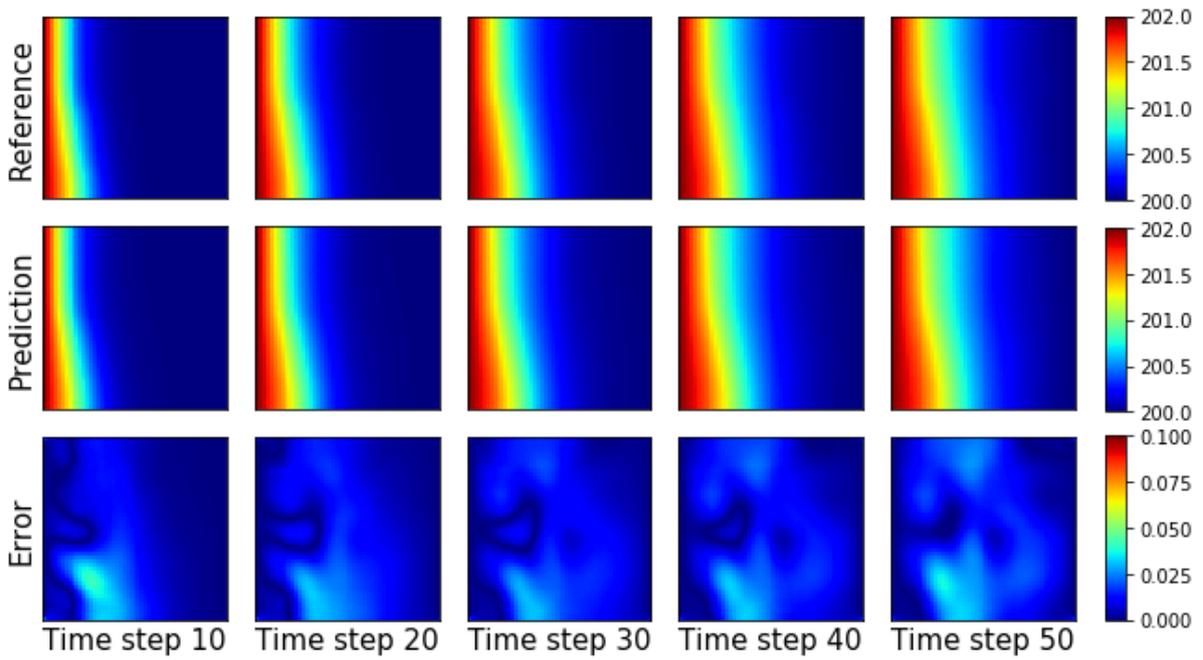

**Figure 19. The hydraulic heads obtained from MODFLOW and the TgAE surrogate for non-Gaussian realization 1.**

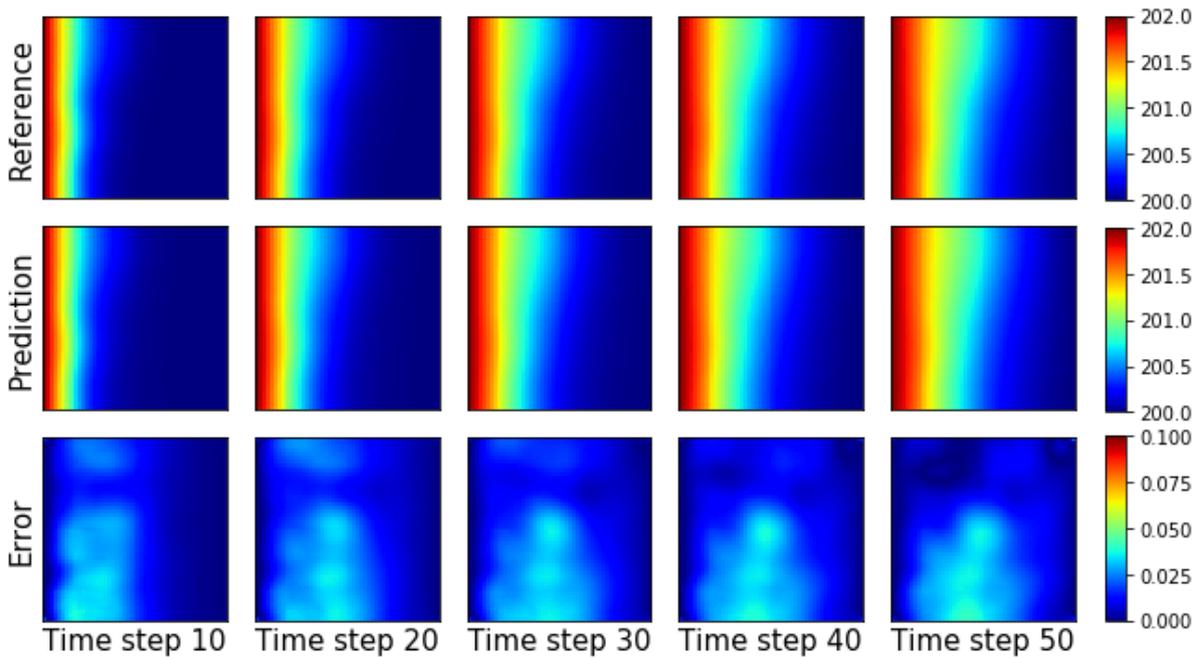

**Figure 20. The hydraulic heads obtained from MODFLOW and the TgAE surrogate for non-Gaussian realization 2.**

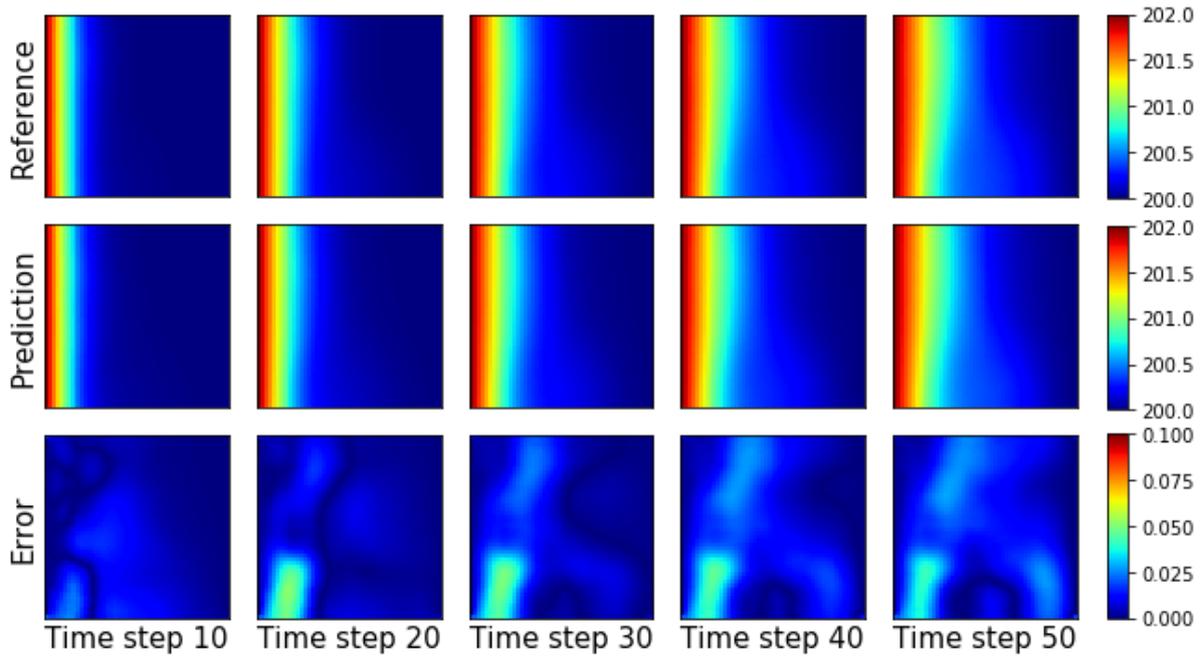

**Figure 21. The hydraulic heads obtained from MODFLOW and the TgAE surrogate for non-Gaussian realization 3.**

### 4.4 Solving inverse problems with TgAE surrogate

In this subsection, the inverse problems of dynamic subsurface flow are solved with the TgAE surrogate based Iterative Ensemble Smoother (IES). Still consider the case introduced in subsection 4.1, and 16 observation points are set in the domain, as shown in **Figure 22**. The hydraulic heads at those points are assumed to be measured and used for inversion of hydraulic conductivity.

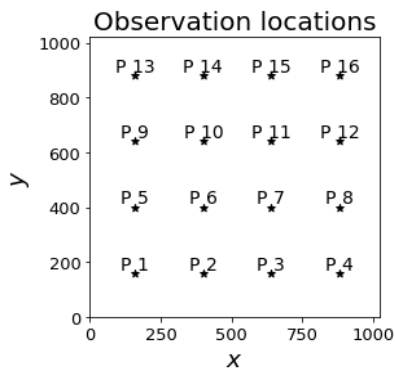

**Figure 22. Observation locations for inverse modeling.**

*4.4.1 Inversion for different correlation length*

Inversion for cases with different correlation length is studied here. First, consider a case with $\eta_x = \eta_y = 408[L]$, and the reference field is shown in **Figure 23(a)**. Then the TgAE surrogate constructed in subsection 4.2 can be used to estimate the hydraulic conductivity field. In the TgAE surrogate based IES for inversion, 100 realizations are used in the ensemble, and we set $\varepsilon_1$=0.01, $\varepsilon_2$=0.0001, and $I_{max}=10$. The initial of log hydraulic conductivity fields and the final mean updated with the TgAE surrogate based IES are shown in **Figure 23(b)** and **(c)**. It can be seen that the major pattern of the reference field can be captured with the available measurements. The initial and final standard deviation of the fields in the ensemble are shown in **Figure 23(d)** and **(e)**, which show that the uncertainty of the stochastic filed is reduced after the inversion. The case with $\eta_x = \eta_y = 204[L]$ is then tested with the proposed method, and the reference field of this case is shown in **Figure 24(a)**. Considering the extrapolation capacity of the TgAE surrogate, the surrogate constructed in subsection 4.2 can also be used for this case. And it is worth mentioning that there is no need to retrain the surrogate for cases with different correlation length, which is an advantage of the TgAE surrogate. The other settings are the same as with the former case, and the inversion results are shown in **Figure 24(b)**, **(c)**, **(d)**, and **(e)**. The general distribution of the reference field has also been captured.

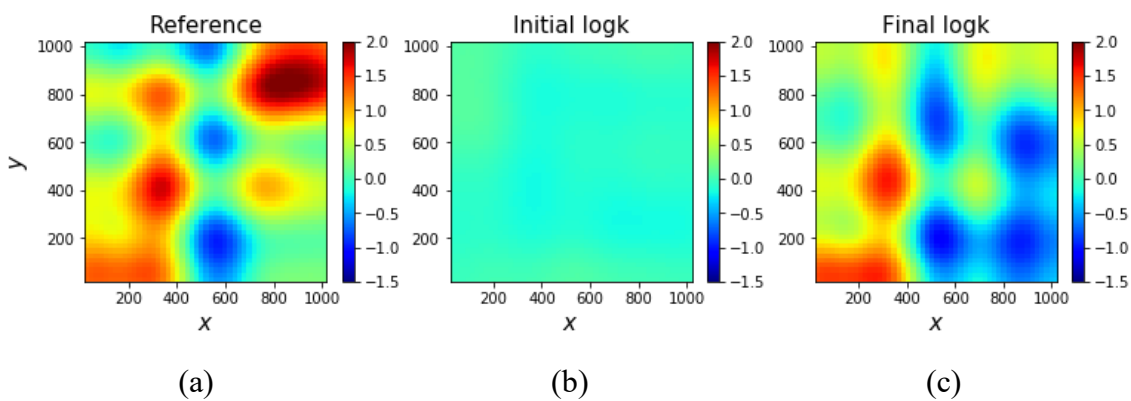

(a)             (b)             (c)

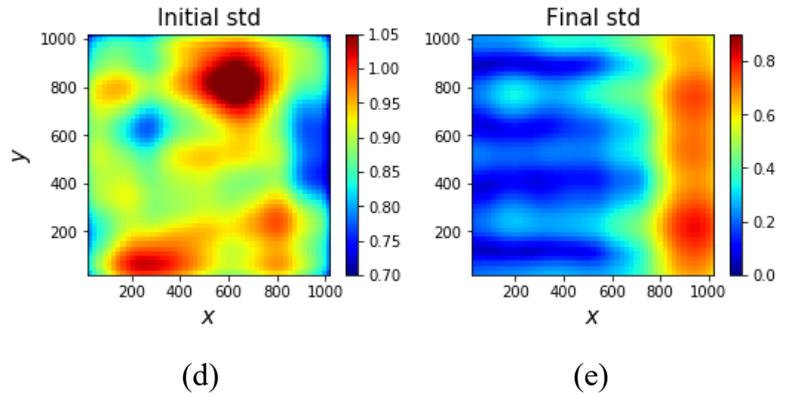

(d)                          (e)

**Figure 23. The reference field for the case with $\eta_x = \eta_y = 408[L]$ (a); Mean of log hydraulic conductivity fields in the ensemble at initial and final step (b), (c); Standard deviation of log hydraulic conductivity fields in the ensemble at initial and final step (d), (e).**

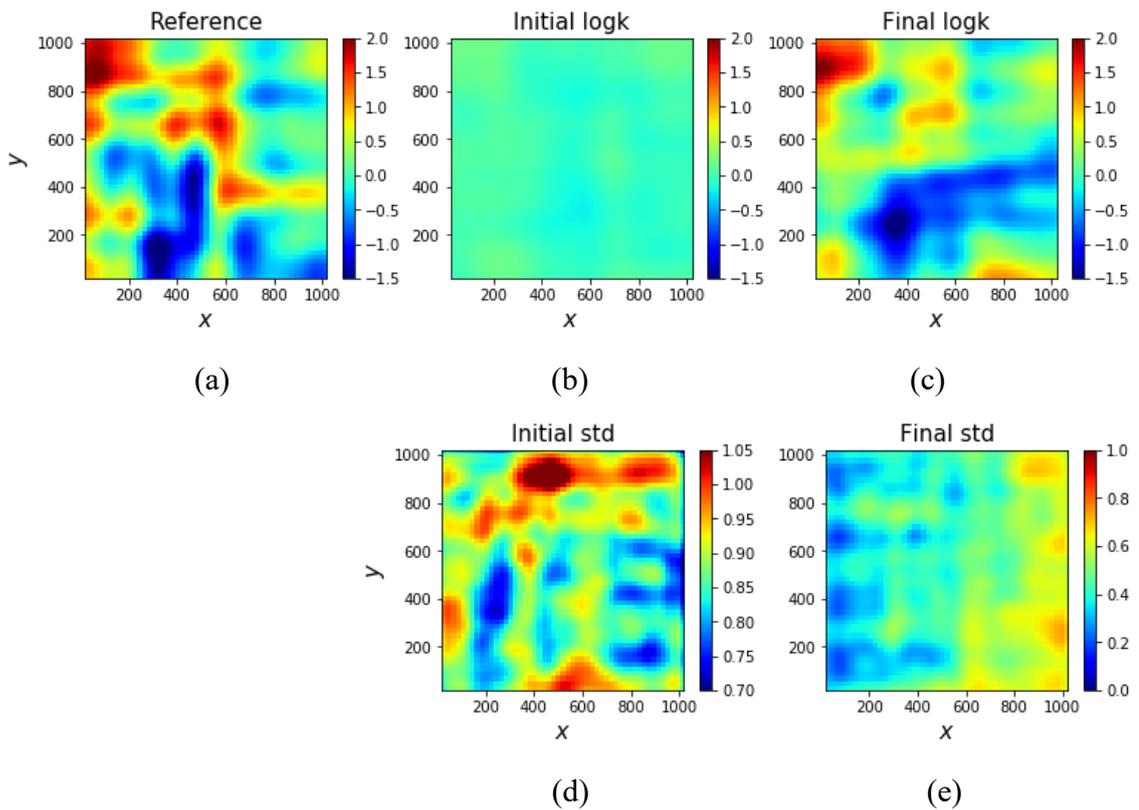

**Figure 24. The reference field for the case with $\eta_x = \eta_y = 204[L]$ (a); Mean of log hydraulic conductivity fields in the ensemble at initial and final step (b), (c); Standard deviation of log hydraulic conductivity fields in the ensemble at initial and final step (d), (e).**

*4.4.2 Inversion for non-Gaussian fields*

The TgAE surrogate based IES is used for inverse modeling of non-Gaussian fields in this subsection. Consider the case introduced in subsection 4.3 with 16 observation locations shown in **Figure 22**, and the constructed TgAE surrogate in subsection 4.3 is used for inversion. The reference field for this case is shown in **Figure 25(a)**, and the other settings for the IES are the same as the former cases. The inversion results of this case are shown in **Figure 25(b)** and **(c)**, and the standard deviation distribution of the realizations in the ensemble before and after the inversion are shown in **Figure 25(d)** and **(e)**. It can be seen that the estimated field is similar to the reference, and the uncertainty of the stochastic field is reduced after the inversion.

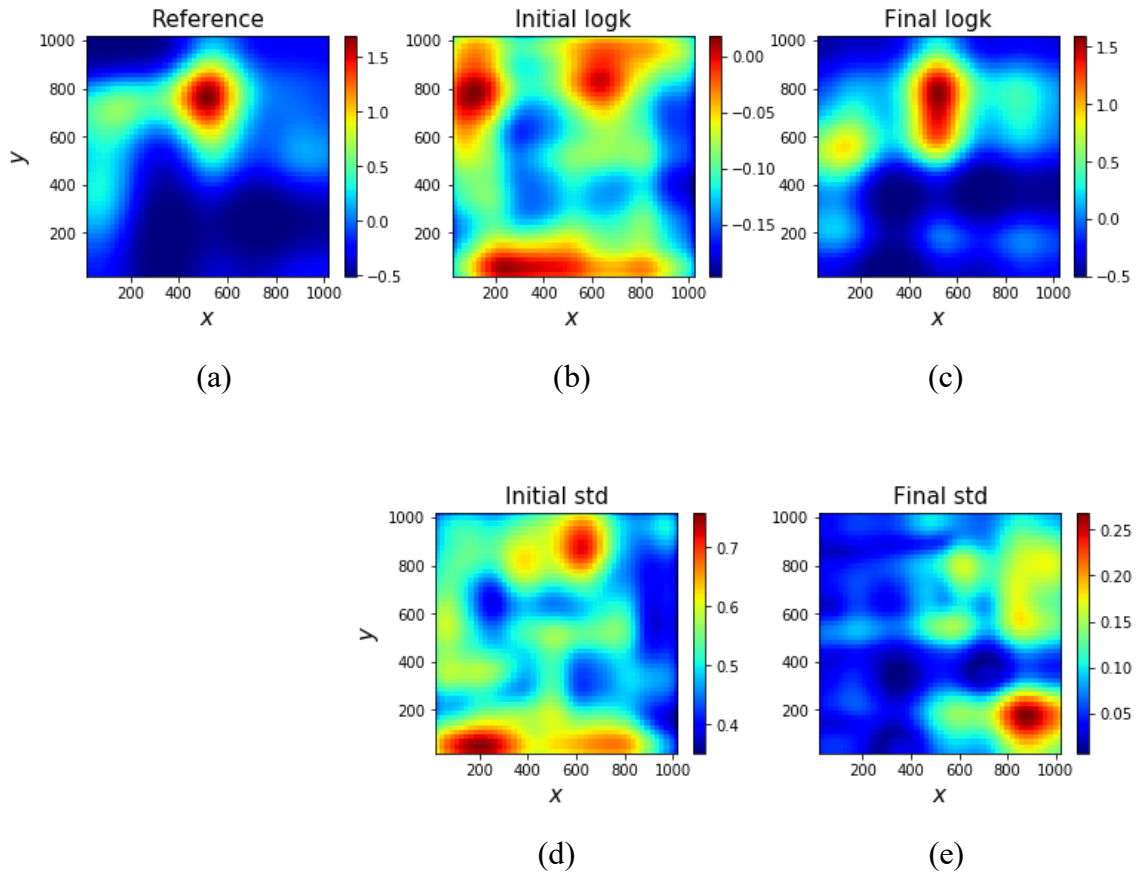

(a)     (b)     (c)

(d)     (e)

**Figure 25. The reference field for the non-Gaussian case (a); Mean of log hydraulic conductivity fields in the ensemble at initial and final step (b), (c); Standard deviation of log hydraulic conductivity fields in the ensemble at initial and final step (d), (e).**

# 5 Discussions and conclusions

In this work, a Theory-guided Auto-Encoder (TgAE) is proposed, which can incorporate the physical constraints into the training of convolutional neural network, and it can be achieved by embedding the finite difference scheme of numerical simulators into the training process. In essence, the solving process of partial differential equations in numerical simulators is replaced by the training process in deep learning, and the solving scheme can be learned by the network. Once trained, the network can be used to predict solutions for cases with different model parameters without the need to run the simulators. And a concept of *virtual realizations* is proposed, which is similar to the collocation points in the PINN framework (Raissi et al., 2019). The proposed TgAE can be used for surrogate modeling and inverse modeling, which can help to improve the efficiency of uncertainty quantification and parameter inversion tasks.

Several subsurface flow cases are introduced to test the performance of the proposed TgAE framework. The results show that the TgAE surrogate can accurately approximate the relationship between the inputs and model responses with limited labeled data. The constructed TgAE surrogate can then be used to quantify the uncertainty of the model responses towards stochastic model parameters. The computational cost of forward evaluation with the TgAE surrogate is much cheaper than running a simulator, so the efficiency of uncertainty quantification is significantly improved. Moreover, the effect of virtual realization numbers towards the accuracy of the TgAE surrogate is studied. It is obvious that adequate virtual realizations can help to guarantee the accuracy of the surrogate when the labeled data are limited. The TgAE surrogate also has some extrapolation ability, which can be used to predict solutions for cases with different correlation lengths and variances. Therefore, there is no need to retrain the surrogate repeatedly while changing some statistical information of the stochastic fields, which is another advantage of the TgAE surrogate. The TgAE is also tested on non-Gaussian fields, which shows satisfactory performance for surrogate modeling.

The constructed TgAE surrogate can also be used for solving the inverse problems. In this work, the Iterative Ensemble Smoother (IES) is combined with the TgAE surrogate to estimate the model parameters with limited observation data. The efficiency of inversion can also be improved without running the simulator iteratively. And once trained, the TgAE surrogate can also be used to solve the cases with different statistical information.

In this work, the deep learning is effectively combined with numerical finite difference methods by constructing the TgAE framework, which shows a new aspect and strategy to merge the data-driven models and physics-based models. In our future work, more complicated problems will be studied with the TgAE framework to expand its application scenarios, for example, the multiphase flow in reservoir simulation.